%% file: main.tex
\documentclass[10pt,twocolumn,letterpaper]{article}

\usepackage{cvpr}                

\usepackage{graphicx}
\usepackage{booktabs}
\usepackage{url}


\definecolor{cvprblue}{rgb}{0.21,0.49,0.74}
\usepackage[pagebackref,breaklinks,colorlinks,allcolors=cvprblue]{hyperref}

\def\paperID{26}
\def\confName{CVPR}
\def\compName{CASTLE}
\def\confYear{2026}

\title{\compName\confYear~Team WDL Technical Report}

\author{
Zhengyang Li, Zhenglin Du, Yi Wen, Fang Liu, Shuo Li, Xu Liu \\
Key Laboratory of Intelligent Perception and Image Understanding \\
{\tt\small 1436210511@qq.com}
}

\begin{document}

\maketitle

\input{sec/0_abstract}
\input{sec/1_Introduction}
\input{sec/2_RelatedWork}
\input{sec/3_Methodology}
\input{sec/4_Experiment}
\input{sec/5_Conclusion}

{
    \small
    \bibliographystyle{ieeenat_fullname}
    \bibliography{reference}
}

\end{document}

%% file: sec/0_abstract.tex
\begin{abstract}
The CASTLE Challenge @ EgoVis 2026 evaluates long-form egocentric video question answering over 600+ hours of multi-perspective recordings. Each four-choice question requires evidence from videos, transcripts, auxiliary photos, people, days, rooms, and temporal context. We propose an evidence-aware multimodal reasoning pipeline based on Qwen. Our system parses question hints, retrieves ASR chunks, attaches auxiliary images, samples candidate video frames, and routes questions into static visual, speech/text, temporal, and mixed types with specialized prompts. Multiple inference passes are aggregated by confidence-weighted voting and converted into the official Codabench format. In ablation, LoRA improves the score from 0.21 to 0.50, and more sampled frames further raise it to 0.58. Our final system ranks first in the CASTLE Challenge @ EgoVis 2026.
\end{abstract}

%% file: sec/1_Introduction.tex
\section{Introduction}
\label{sec:introduction}

Egocentric video understanding requires reasoning about long-term human behavior, object states, conversations, and spatial layouts from first-person and environmental cameras. The CASTLE Challenge @ EgoVis 2026 focuses on closed-form question answering: given the entire CASTLE dataset, the system must select the correct option from four candidate answers for each query. This setting is substantially different from short-clip video QA because the evidence for a question may appear in a small moment within hundreds of hours of recordings and may be distributed across multiple camera streams or transcript files.

The main difficulty is therefore not only visual recognition, but also evidence localization. Some questions are static visual questions, such as counting objects, reading visible text, identifying colors, or locating an item in a room. Others depend primarily on speech transcripts, such as quiz topics, named entities, jokes, or stated numbers. A single prompt or a single sampled frame strategy is insufficient for these heterogeneous cases.

Our contributions are summarized as follows:
\begin{itemize}
    \item We build an official-data-only evidence construction pipeline for long-form CASTLE video QA, combining transcript retrieval, image discovery, and frame sampling.
    \item We introduce question-type-aware prompting for static visual, speech/text, temporal, and mixed questions.
    \item We use multi-sample self-consistency with strict, balanced, and aggressive prompt variants to improve answer reliability.
    \item We report an ablation path showing that LoRA adaptation and higher frame sampling substantially improve the challenge score.
\end{itemize}

%% file: sec/2_RelatedWork.tex
\section{Related Work}
\label{sec:relatedwork}

\subsection{Egocentric Video Question Answering}
Egocentric video QA aims to answer natural-language questions from first-person or multi-perspective videos. Compared with standard image QA, egocentric QA requires modeling object manipulation, social interaction, temporal order, and viewpoint changes. CASTLE provides multi-perspective egocentric videos for multimodal understanding~\cite{castle2025dataset}, and the CASTLE@EgoVis challenge further increases the difficulty by requiring retrieval over long recordings rather than reasoning from a short pre-trimmed clip~\cite{castle2026challenge}.

\subsection{Multimodal Large Language Models}
Recent multimodal large language models can jointly process text and visual evidence, making them strong candidates for video QA. TimeChat studies time-sensitive multimodal modeling for long video understanding~\cite{ren2024timechat}, while Qwen-based models provide strong general reasoning ability~\cite{qwen2025}. However, directly feeding a small number of frames to an MLLM is often unreliable for long-form datasets, because the relevant evidence may be missing from the sampled frames.

\subsection{Retrieval-Augmented Reasoning}
Retrieval-augmented generation improves factual grounding by adding external evidence to the model context. In this challenge, retrieval is performed over official ASR transcripts, auxiliary photos, and video frames. Similar fine-grained visual interpretation problems have also been studied in surgical video understanding~\cite{nwoye2022rendezvous} and remote sensing interpretation~\cite{yan2026language,chai2026like}. Unlike generic vector retrieval, our pipeline uses task-specific hints such as day, person, room, option words, and temporal keywords to rank candidate evidence.

\subsection{Parameter-Efficient Adaptation}
LoRA is a parameter-efficient fine-tuning method that updates low-rank adapter matrices while keeping most model parameters frozen~\cite{hu2022lora}. In our experiments, LoRA adaptation is used to improve the base model's ability to follow the CASTLE multiple-choice QA format and to better align answer prediction with competition-style queries.

%% file: sec/3_Methodology.tex
\section{Methodology}
\label{sec:methodology}

\begin{figure*}[t]
    \centering
    \includegraphics[width=0.92\textwidth]{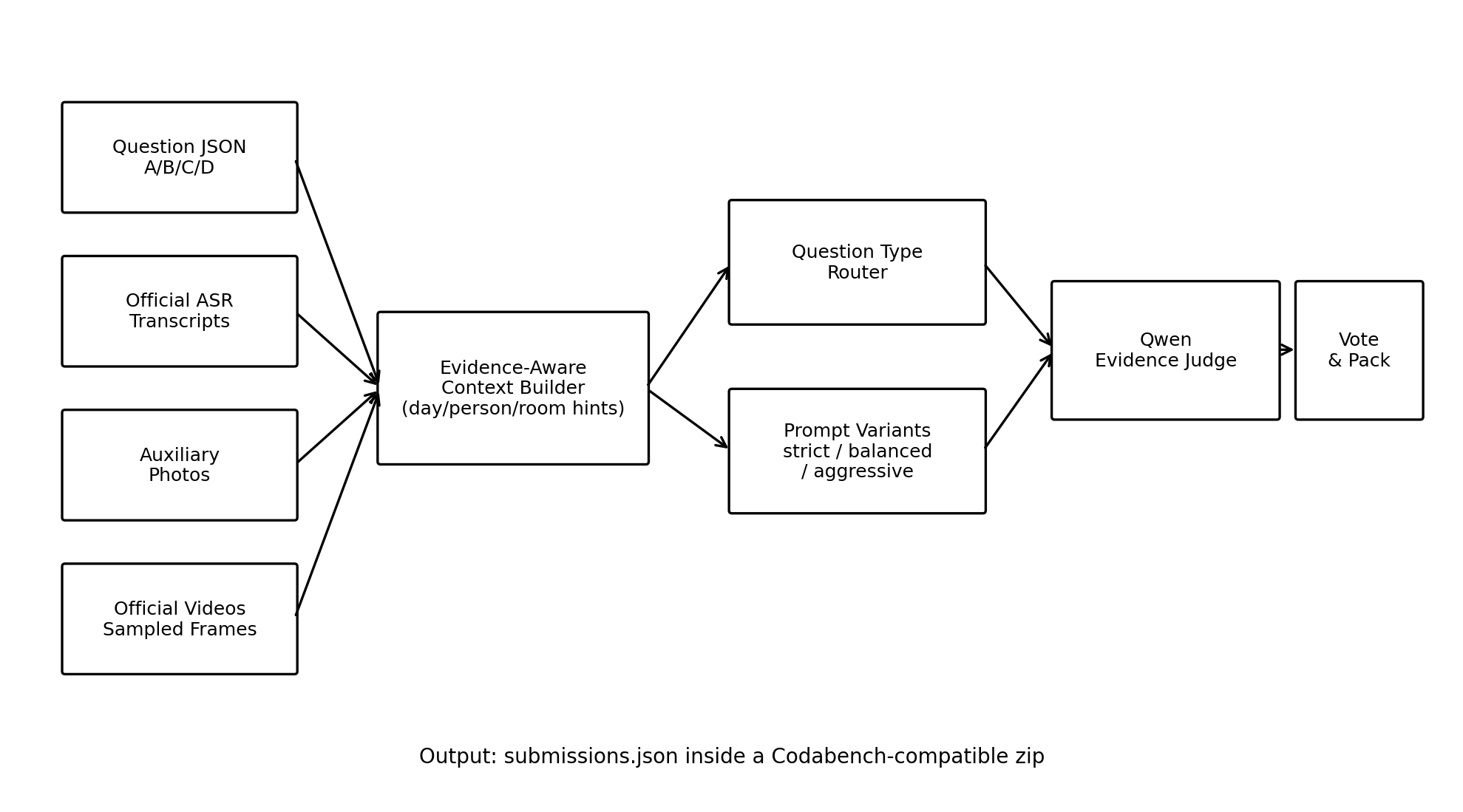}
    \caption{Overview of the proposed CASTLE question answering pipeline. The system builds question-specific evidence from official artifacts, routes the query to specialized prompts, performs multiple Qwen inference passes, and produces a Codabench-compatible submission package.}
    \label{fig:pipeline}
\end{figure*}

Figure~\ref{fig:pipeline} shows the overall framework. Our system contains four major components: evidence-aware context construction, question-type routing, Qwen-based evidence judging, and confidence-weighted voting.

\subsection{Task Formulation}
Given a question $q$ and four answer candidates $\{a,b,c,d\}$, the goal is to predict a single option $\hat{y}\in\{a,b,c,d\}$. Unlike short-video QA, the input evidence is not a single trimmed clip. Instead, the possible evidence space includes long videos, multi-perspective camera streams, ASR transcripts, and auxiliary images. Therefore, we formulate the task as evidence selection followed by answer classification:
\begin{equation}
    \hat{y}=\arg\max_{y\in\{a,b,c,d\}} S(y\mid q, E_q),
\end{equation}
where $E_q$ denotes the retrieved context for question $q$, and $S(\cdot)$ is the confidence score produced by the Qwen evidence judge and voting module.

\subsection{Evidence-Aware Context Construction}
For each question, the system first parses explicit and implicit hints from the query. We detect person names, day expressions, room names, visual keywords, speech-related keywords, and temporal keywords. These hints guide retrieval from official data sources.

\paragraph{Transcript retrieval.}
Official ASR transcripts are divided into timestamped chunks. Each chunk is assigned a score according to lexical overlap with the question, overlap with answer options, phrase-level matches, and contextual bonuses for matched day, person, or room. We keep the top-$K$ transcript chunks and include their timestamp, source path, speaker or camera entity, and text content in the prompt. This design prevents the model from choosing an answer merely because an option word appears in an unrelated transcript segment.

\paragraph{Auxiliary image retrieval.}
Auxiliary photos are ranked using path-level and keyword-level hints. If a query mentions a person, room, day, object, or visible attribute, image paths matching those hints are prioritized. This component is particularly useful for static visual questions involving color, object identity, room layout, brand, logo, count, or OCR-like text.

\paragraph{Video frame sampling.}
For visual and temporal questions, the system selects candidate videos according to day, room, and person hints. From each matched video, frames are uniformly sampled and resized when necessary. Increasing the number of sampled frames enlarges the visual coverage and reduces the probability that the decisive evidence is missed.

\subsection{Question-Type Routing and Prompt Design}
The query is routed into one of four types: static visual, speech/text, temporal, or mixed. Each type uses a different instruction block.

Static visual prompts ask the model to prioritize images and video frames, inspect object counts, colors, visible text, brands, spatial relations, and room locations. Speech/text prompts prioritize timestamped transcript evidence and require the model to match the correct speaker, day, and event before accepting an option. Temporal prompts ask the model to reconstruct order from timestamps and clip numbers, especially for first, last, before, after, while, and during questions. Mixed prompts combine transcript and visual evidence while requiring consistency with the question condition.

\begin{figure}[t]
    \centering
    \includegraphics[width=0.48\textwidth]{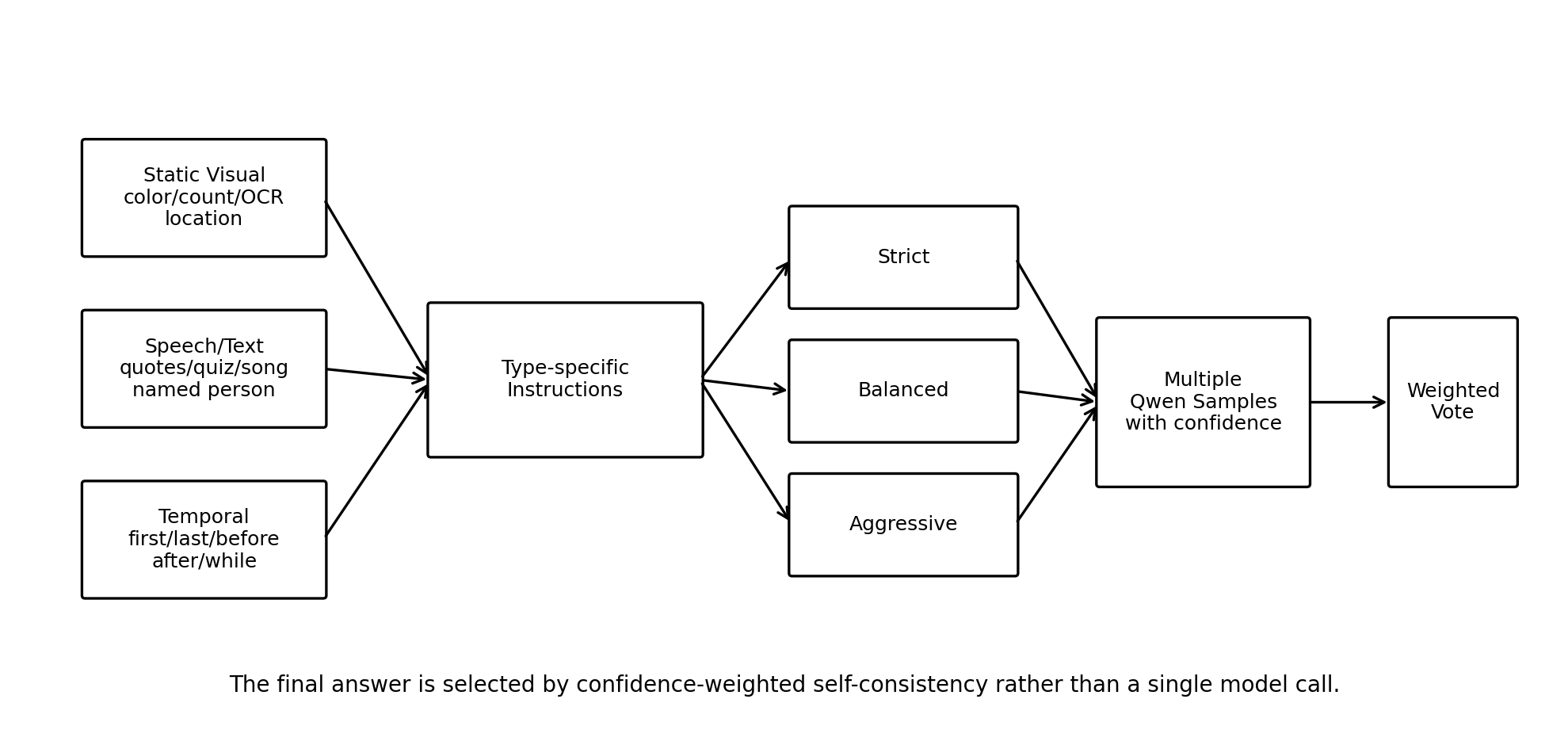}
    \caption{Question-type-aware prompting and confidence-weighted voting. Each question is answered multiple times using strict, balanced, and aggressive policies, then aggregated into a final answer.}
    \label{fig:voting}
\end{figure}

\subsection{LoRA Adaptation}
In addition to the retrieval-based inference pipeline, we evaluate LoRA adaptation to align the base model with the CASTLE multiple-choice QA format. During LoRA fine-tuning, the base model parameters are frozen and trainable low-rank matrices are inserted into selected linear projection layers. The model is optimized to generate only the final answer option. The loss is computed on answer tokens:
\begin{equation}
    \mathcal{L}=-\sum_{j\in\mathcal{A}}\log P(x_j\mid x_{<j}, q, E_q),
\end{equation}
where $\mathcal{A}$ denotes the answer-token positions. This adaptation improves instruction following and reduces invalid or over-explanatory outputs.

\subsection{Multi-Sample Qwen Evidence Judging}
The system calls Qwen multiple times for each context. We use three complementary prompt variants. The strict variant accepts only directly supported evidence whenever possible. The balanced variant prefers direct evidence and then grounded reasoning from nearby context. The aggressive variant makes a decisive choice even when the evidence is incomplete. Each response is required to return a JSON object containing the answer, confidence, evidence type, rationale, and key evidence.

\subsection{Confidence-Weighted Voting and Submission Packing}
Let $r_i$ denote one model response with answer $y_i$ and confidence $c_i$. The final score of option $y$ is computed as
\begin{equation}
    V(y)=\sum_i \mathbb{I}(y_i=y)\cdot c_i\cdot w_i,
\end{equation}
where $w_i$ is a mode-specific weight, such as a higher weight for vision-mode responses when visual evidence is attached. The final prediction is the option with the highest aggregated score. We generate two submission packages: a strict package requiring stronger agreement and an attack package accepting lower agreement to maximize coverage. The final output is packed as a \texttt{submissions.json} file inside a zip archive following the Codabench format.

%% file: sec/4_Experiment.tex
\section{Experiment}
\label{sec:experiment}

\subsection{Experimental Setup}

\paragraph{Data.}
We use only official/local competition artifacts, including the CASTLE question JSON, official ASR transcripts, official auxiliary photos, and locally downloaded official video subsets. The submitted file follows the official closed-form format, where every question id is mapped to one answer letter among \texttt{a}, \texttt{b}, \texttt{c}, and \texttt{d}.

\paragraph{Context construction.}
For the final high-cost run, we build rich contexts with 30 transcript chunks, up to 16 auxiliary images, video-frame extraction enabled, 32 frames per video, and at most 4 candidate videos per question. These values provide a trade-off between evidence coverage and API/runtime cost.

\paragraph{Inference.}
The default model is \texttt{qwen3-max} in text mode. Vision-mode inference can be enabled with \texttt{qwen-vl-max-latest}. For each question, we run multiple samples using strict, balanced, and aggressive prompt variants. The default temperature is set to 0.18 in the final script. Predictions are saved as JSONL files and then aggregated by the voting module.

\paragraph{Evaluation metric.}
We report the Codabench leaderboard score/accuracy. Since the task is a four-choice QA benchmark, the primary measure is the proportion of correctly selected options.

\subsection{Main Results and Ablation Study}
Table~\ref{tab:ablation} summarizes the main improvement path. The baseline obtains a score of 0.21. LoRA fine-tuning significantly improves the score to 0.50, indicating that adapting the model to the CASTLE multiple-choice QA style is important. Increasing the number of sampled frames further improves the score to 0.58 by providing broader visual coverage for evidence localization.

\begin{table}[t]
    \centering
    \caption{Ablation study on the CASTLE Challenge @ EgoVis 2026.}
    \label{tab:ablation}
    \begin{tabular}{lcc}
        \toprule
        Method & Score & Gain \\
        \midrule
        Baseline Qwen inference & 0.21 & -- \\
        + LoRA fine-tuning & 0.50 & +0.29 \\
        + More sampled video frames & 0.58 & +0.08 \\
        \bottomrule
    \end{tabular}
\end{table}

\subsection{Analysis}
The result suggests that the main bottleneck of CASTLE is evidence coverage. LoRA improves the model's answer format and domain alignment, but many errors still occur when the decisive visual or transcript evidence is absent from the context. Increasing the sampled frame count improves performance because more candidate moments are visible to the model. The confidence-weighted voting stage further stabilizes predictions by reducing the influence of a single uncertain model call.

\subsection{Implementation Details}
The pipeline is implemented as a command-line Python script with three main stages. First, \texttt{build-context} constructs JSONL contexts from questions and local evidence. Second, \texttt{infer} calls Qwen with the selected mode, model, sampling count, and prompt variants. Third, \texttt{vote-pack} aggregates all predictions and writes a Codabench-compatible zip file. A separate \texttt{validate} command checks missing, extra, or invalid answers before submission.

%% file: sec/5_Conclusion.tex
\section{Conclusion}
\label{sec:conclusion}
We present an evidence-aware Qwen system for the CASTLE Challenge @ EgoVis 2026. It turns long-form multi-perspective video QA into structured evidence retrieval and verification. Using transcript retrieval, auxiliary images, video-frame sampling, question-specific prompts, LoRA, and confidence-weighted voting, our system achieves strong results on the CASTLE four-choice QA task. Future work includes better temporal localization, cross-camera evidence fusion, and automatic strategy calibration.